%% file: iclr2027_conference.tex
\documentclass{article} %
\usepackage{iclr2027_conference,times}

\input{math_commands.tex}

\usepackage{hyperref}
\usepackage{url}
\usepackage{graphicx}
\usepackage{caption}
\usepackage{xspace}
\usepackage{xcolor}
\usepackage{tabularx}
\usepackage{booktabs}
\usepackage{multirow}
\usepackage{colortbl}
\usepackage{wrapfig}
\usepackage{needspace}
\usepackage{placeins}
\usepackage{float}

\newcommand{\algname}{\textit{Graph3R}\xspace}

\newcommand{\tb}[1]{\textbf{#1}}
\newcommand{\cornerarrow}{\ensuremath{\hookrightarrow}\,}

\title{
Graph3R: Unifying View Graph Initialization with Feed-Forward Reconstruction
}

\usepackage{etoolbox}
\makeatletter
\patchcmd{\@maketitle}
  {\begin{tabular}[t]{l}\bf\rule{\z@}{24pt}\@author\end{tabular}}
  {\makebox[\textwidth][c]{\begin{tabular}[t]{c}\bf\rule{\z@}{24pt}\@author\end{tabular}}}
  {}{\PackageError{author-layout}{Could not center author block}{}}
\makeatother
\author{%
Jingwen Yu$^{1,2}$ \quad
Hengli Wang$^{1}$ \quad
Jianhao Jiao$^{3}$ \\[0.2em]
\textbf{Zhonghang Liu}$^{4}$ \quad
\textbf{Ping Tan}$^{1}$ \quad
\textbf{Hong Zhang}$^{2}$\\[0.5em]
{\normalfont\small $^{1}$Hong Kong University of Science and Technology}\\[0.1em]
{\normalfont\small $^{2}$Southern University of Science and Technology}\\[0.1em]
{\normalfont\small $^{3}$China Merchants Research Institute of Advanced Technology}\\[0.1em]
{\normalfont\small $^{4}$SmartMore Ltd.}
}

\iclrfinalcopy %

\begin{document}

\maketitle
\lhead{Preprint}
\vspace{-0.49cm}

\begin{figure}[htbp]
    \centering
    \includegraphics[width=0.92\linewidth]{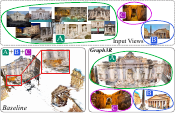}
    \caption{
    \textbf{Comparison with existing baselines}, given input views from three disjoint scenes (A--C), the baseline method ${\pi}^{3}$~\citep{wang2025pi3}
    produces an incorrect reconstruction (left).
    While ours (\algname) predicts a view graph from shared geometric features, partitions
    the views into scene groups, and reconstructs each group separately using
    a single backbone pass (right).
    }
    \label{fig:teaser}
\end{figure}

\begin{abstract}
Feed-forward reconstruction recovers camera poses and dense geometry from image sets whose views are assumed to come from a single scene.
In practice, uncurated image collections often contain views from disjoint scenes, and reconstructing these views jointly can corrupt camera poses and geometry.
Existing pipelines typically resolve view associations in separate stages before final reconstruction, leaving view graph initialization and reconstruction inference only loosely coupled.
We introduce \algname, which \textbf{\textit{unifies view graph initialization with feed-forward reconstruction through a shared geometric representation}}.
A lightweight verification head predicts pairwise covisibility from the multi-view features of $\pi^3$ to construct a view graph.
This graph classifies the input views into scene groups.
For each group, we mask the shared features before decoding camera poses and dense geometry.
Reusing the multi-view features allows \algname to reconstruct disjoint scenes separately with a single backbone pass.
Across three benchmarks, \algname improves average maximum recall at 100\% precision by upto 8.99 percentage points over the baseline, while improving reconstruction quality on mixed-scene inputs.
Regarding the multi-session mapping task, \algname improves cross-session localization and achieves competitive pose estimation accuracy, while delivering a runtime speedup of approximately $35.9\times$ over the baseline when reconstructing the query group.
By integrating learned view association with geometric inference, our approach takes a step toward end-to-end reconstruction from uncurated image collections.
\end{abstract}

\input{sections/intro}
\input{sections/related_work.tex}
\input{sections/method.tex}

\input{sections/exp.tex}

\input{sections/conclusion.tex}

\bibliography{iclr2027_conference}
\bibliographystyle{iclr2027_conference}

\input{appendix}
\end{document}

%% file: math_commands.tex
\usepackage{amsmath,amsfonts,bm}

\def\eqref#1{equation~\ref{#1}}

\def\1{\bm{1}}

\DeclareMathAlphabet{\mathsfit}{\encodingdefault}{\sfdefault}{m}{sl}
\SetMathAlphabet{\mathsfit}{bold}{\encodingdefault}{\sfdefault}{bx}{n}

%% file: sections/intro.tex
\input{figures/pipeline.tex}
\section{Introduction} \label{sec:intro}
Feed-forward 3D reconstruction methods such as MASt3R~\citep{leroy2024grounding}, VGGT~\citep{wang2025vggt}, and $\pi^3$~\citep{wang2025pi3} enable efficient recovery of camera poses and dense geometry from unordered images.
These methods typically assume that the input views form a geometrically connected scene.
Uncurated image collections, including Internet photos and images gathered across robot sessions, can violate this assumption by mixing views from disjoint scenes.
Processing these views jointly can corrupt both camera poses and scene geometry~\citep{han2025emergent}, as illustrated in Fig.~\ref{fig:teaser}.
Reliable reconstruction therefore depends on identifying which views should be reconstructed together.

Classical structure-from-motion (SfM)~\citep{schoenberger2016mvs, schoenberger2016sfm} and simultaneous localization and mapping (SLAM)~\citep{campos2021orb, mur2015orb} systems establish view associations through image retrieval and geometric verification.
These associations define a \tb{view graph}, whose nodes represent images and whose edges indicate covisibility.
Visual place recognition (VPR) methods~\citep{Berton_2025_MegaLoc} propose candidate pairs, and geometric verification checks the spatial consistency of local feature correspondences using robust estimation such as RANSAC.
However, correspondence consistency alone does not guarantee covisibility.
GV-Bench~\citep{yu2024gv} and Doppelgangers~\citep{cai2023doppelgangers} document verification failures under long-term appearance changes and perceptual aliasing, including false associations between visually similar but distinct places.

Fig.~\ref{fig:paradigms} contrasts how representative pipelines connect view association to reconstruction.
GLUEMAP~\citep{pan2026gluemap} builds a view graph through image retrieval and pairwise verification before applying a feed-forward model to local view neighborhoods.
RobustVGGT~\citep{han2025emergent} identifies outlier views from the reconstruction model's internal representations, then reruns the backbone on the retained inputs.
These pipelines require separate feature computation for verification or a second backbone pass for reconstruction.
Doppelgangers++~\citep{xiangli2025doppelgangers} already shares MASt3R features for graph construction and pruning within an SfM pipeline, demonstrating the value of feature reuse.
This motivates a further question: \textbf{\textit{Can view graph initialization and feed-forward reconstruction be unified through one shared backbone pass?}}

We introduce \algname, a $\pi^3$-based model that unifies learned view graph initialization and feed-forward reconstruction through a shared geometric representation.
A lightweight verification head predicts pairwise covisibility from intermediate multi-view features.
The resulting view graph partitions the input into scene groups, whose masks condition pose and geometry decoding.
The backbone features are computed once and reused to reconstruct each group, allowing disjoint scenes to be recovered separately without a second backbone pass.

Our contributions are threefold:
(a) We unify learned view graph initialization and feed-forward reconstruction through shared backbone features, avoiding repeated feature extraction between the two tasks.
(b) We develop graph-conditioned decoding that masks latent embeddings by inferred scene group to recover camera poses and dense geometry separately for disjoint scenes.
(c) We evaluate verification accuracy, reconstruction quality, and inference efficiency on uncurated image collections, and demonstrate \algname in a real-world multi-session mapping system.

%% file: figures/pipeline.tex
\begin{figure}[ht]
    \centering
    \includegraphics[width=0.98\linewidth]{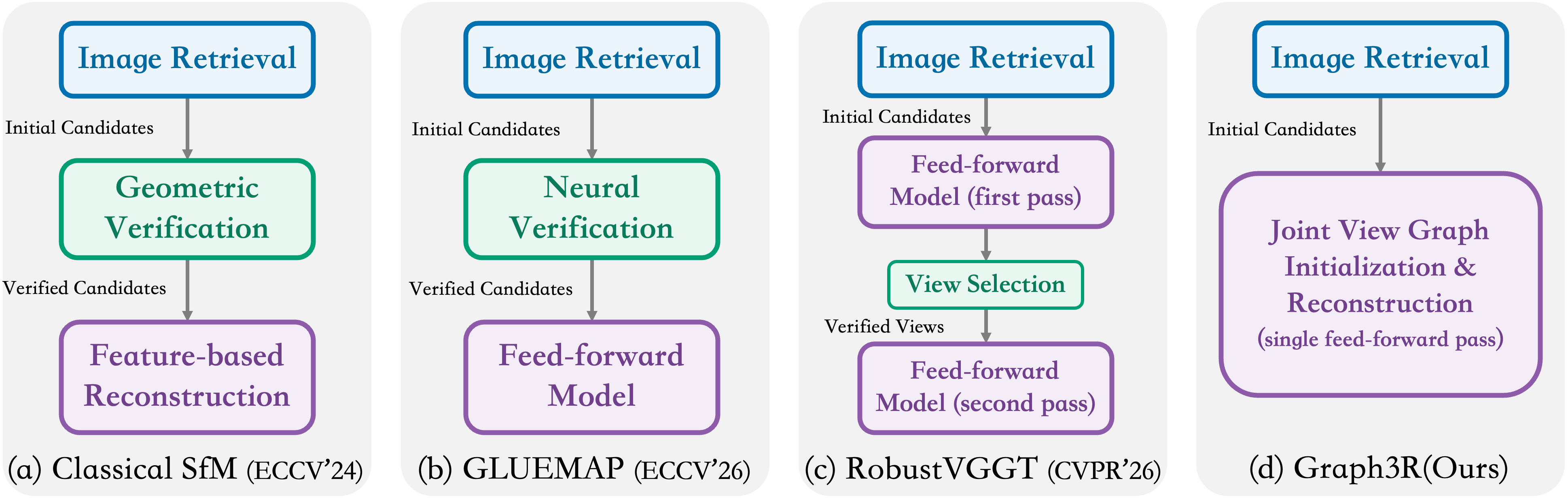}
    \caption{\textbf{Comparison of different SfM paradigms.}
    Classical SfM~\cite{pan2024global} and GLUEMAP~\citep{pan2026gluemap} initialize a view graph through separate verification stages before reconstruction.
    RobustVGGT~\citep{han2025emergent} selects views using VGGT's latent features and reruns the feed forward model on the selected views.
    \tb{\algname} proposes a unified paradigm that jointly predicts the view graph and scene structure with a single backbone inference as detailed illustrated in Fig.~\ref{fig:method}.
    }
    \label{fig:paradigms}
\end{figure}

%% file: sections/related_work.tex
\section{Related Work} \label{sec:relatedworks}

\subsection{Feed-forward Reconstruction and Structure-from-Motion} \label{related-feedforward}

Feed-forward reconstruction models recover scene geometry through learned multi-view representations.
MASt3R~\citep{leroy2024grounding} combines pairwise pointmap prediction with learned descriptors for dense matching, while VGGT~\citep{wang2025vggt} jointly predicts camera parameters and dense geometry from multiple views.
\mbox{$\pi^3$}~\citep{wang2025pi3} introduces a permutation-equivariant architecture that removes the need for a designated reference view.
These advances improve geometric inference, but reconstruction remains sensitive to input connectivity: unrelated views can corrupt both camera poses and scene geometry~\citep{han2025emergent}.

GLUEMAP~\citep{pan2026gluemap} combines feed-forward local reconstruction with global SfM optimization.
It retrieves and verifies image pairs to construct a sparse view graph, applies $\pi^3$ to local graph neighborhoods, and aligns and refines the resulting reconstructions globally.
The graph restricts local inference to selected views, but its initialization requires verification features computed separately from the reconstruction backbone.
\algname integrates graph prediction into the reconstruction model and reuses the same backbone features for decoding each inferred scene group.

\subsection{View Graph Initialization} \label{related-verification}

View graph initialization depends on distinguishing covisible views from visually similar distractors.
VPR methods~\citep{keetha2023anyloc, ye2023condition, Berton_2025_MegaLoc} retrieve candidate views using global descriptors, while local matchers~\citep{edstedt2025romav2} provide correspondences for geometric verification.
GV-Bench~\citep{yu2024gv} evaluates the reliability of this verification step under long-term appearance changes and perceptual aliasing, exposing limitations of decisions based on geometric inliers.

Learned verification methods address visual ambiguity explicitly.
Doppelgangers~\citep{cai2023doppelgangers} classifies image pairs using appearance and correspondence cues, and Doppelgangers++~\citep{xiangli2025doppelgangers} builds a classifier on MASt3R's geometric features.
Its integration with MASt3R-SfM shares features for graph construction and pruning, with the verified graph guiding subsequent SfM reconstruction.
OpenNavMap~\citep{jiao2026opennavmap} leverages the learned confidence map of MASt3R to reject outlier matches while it only focuses on pixel-level verification and does not provide a global view association.
RobustVGGT~\citep{han2025emergent} provides a training-free alternative: it selects relevant views using internal attention scores or feature similarities, then reruns the backbone on the retained inputs.

\algname predicts pairwise covisibility with a learned verification head operating on shared multi-view features.
The predicted graph conditions group-wise pose and geometry decoding from these retained features, coupling view association and reconstruction within one backbone pass.
Table~\ref{tab:related-works} in Appendix~\ref{app:paradigm-comparison} compares the association features, decision rules, and reconstruction stages of representative methods.

\subsection{Multi-session Mapping} \label{related-mapping}
Multi-session mapping relies on reliable associations across sessions to align and merge maps. A typical pipeline retrieves candidate views from existing sessions and filters outliers using a learned verifier such as Doppelgangers++~\citep{xiangli2025doppelgangers} or geometric verification~\citep{yu2024gv}, before estimating relative poses.
MASt3R-SLAM~\citep{murai2025mast3r} and VGGT-SLAM~\citep{maggio2026vggt} combine feed-forward geometry with global map optimization for single-session mapping, while FLOCC-SLAM~\citep{lajoie20253d} and OpenNavMap~\citep{jiao2026opennavmap} apply feed-forward model for relative pose estimation across sessions.
Specifically, FLOCC-SLAM uses these models for inter-robot loop closing.
OpenNavMap~\citep{jiao2026opennavmap} utilizes the calibrated confidence map to reject outlier matches.
MR.ScaleMaster~\citep{ju2026mrscalemaster} addresses inter-session scale ambiguity through a $\mathrm{Sim}(3)$ pose-graph backend.
\algname complements these systems by integrating view association into local reconstruction.
Given images from disjoint scenes, it predicts a view graph and recovers camera poses and geometry separately for each inferred scene group.
The resulting reconstructions support downstream alignment and map refinement.

%% file: sections/method.tex
\section{Method}
\label{sec:method}

\subsection{Problem Formulation and System Overview}
\label{sec:system-overview}
\input{figures/system_pipeline.tex}
\textbf{Problem Formulation.}
Given an image collection $\mathcal{I}=\{\mathbf{I}_i\}_{i=1}^{N}$, our goal is to identify groups of views that should be reconstructed together and recover the corresponding camera poses and dense geometry.
The collection may contain views from several disjoint scenes, and the number of scenes is unknown.
We follow the classical SfM pipeline~\citep{pan2024global} and formulate the problem as a two-stage process: \tb{(a) view graph initialization}, which predicts pairwise covisibility between images and constructs a view graph $\mathcal{G}=(\mathcal{V},\mathcal{E})$, where each vertex corresponds to an image and an edge indicates predicted geometric overlap, and \tb{(b) graph-conditioned reconstruction}, which selects shared latent features by scene group and reconstructs each group separately.

\textbf{System Overview.}
As illustrated in Fig.~\ref{fig:method}, \tb{\algname} builds on $\pi^3$~\citep{wang2025pi3} and unifies the view verification and reconstruction with the geometric representation of the shared feature aggreation backbone.
We denote the shared feature aggregation backbone by $\mathcal{B}_{\theta}$, which takes the input image collection $\mathcal{I}$ and produces shared features $\mathbf{Z}=\mathcal{B}_{\theta}(\mathcal{I})\in\mathbb{R}^{N\times P\times D}$, where $N$ is the number of views, $P$ is the number of tokens per view, and $D$ is the concatenated feature dimension.
A verification head (described in Sec.~\ref{sec:method-verification-head}) predicts pairwise covisibility from $\mathbf{Z}$ and constructs a view graph $\mathcal{G}$, whose connected components define the scene groups.
For each group, we gather the corresponding view features from $\mathbf{Z}$ and pass them together to the geometry decoder, achieving graph-conditioned reconstruction (Sec.~\ref{sec:method_graph-conditioned}).

\subsection{Learned View Graph Prediction}
\label{sec:method-verification-head}
\input{figures/verification_head.tex}
\paragraph{Pairwise covisibility prediction.}
We adpot a similar design of the $\pi^3$'s camera decoder for the verification head, with an additional prediction head to produce covisibility scores as illustrated in Fig.~\ref{fig:verification-head-architecture}.
In specific, the verification head takes the shared feature representation $\mathbf{Z}$ as input and processes it with a stack of transformer decoder blocks, followed by an MLP and pooling layer to generate a per-view latent descriptor $\mathbf{v}_i\in\mathbb{R}^{C}$.
The pairwise prediction head $h_{\phi}$ maps each descriptor pair to a scalar logit, which is converted to a covisibility probability by a sigmoid:
\begin{equation}
    S_{ij}=\sigma(h_{\phi}(\mathbf{v}_i,\mathbf{v}_j)),
    \qquad \mathbf{S}\in[0,1]^{N\times N},
    \label{eq:method-pair-scores}
\end{equation}
where $\sigma$ is the sigmoid function and $\mathbf{S}$ is the pairwise covisibility score matrix.
Here, $S_{ij}$ estimates whether views $i$ and $j$ overlap geometrically, rather than merely belonging to the same scene.

\paragraph{View graph construction and scene grouping.}
We threshold the pairwise scores at $\tau_s$ to construct an undirected view graph $\mathcal{G}=(\mathcal{V},\mathcal{E})$, identifying vertices with view indices $\mathcal{V}=\{1,\ldots,N\}$.
Let $\mathcal{V}_1,\ldots,\mathcal{V}_K\subseteq\mathcal{V}$ denote the vertex sets of its connected components, where $K$ is the number of predicted components, including isolated vertices.
The threshold determines accepted pairwise links, while connectivity determines the groups passed to reconstruction.
Consequently, two views can belong to the same group through a path of overlapping views even when their direct score is below $\tau_s$.
For multi-view reconstruction, views with no accepted links to other images are treated as isolated inputs and excluded from graph-conditioned decoding.

\subsection{Graph-Conditioned Reconstruction}
\label{sec:method_graph-conditioned}

The predicted view graph determines which views are decoded together.
For each scene group $\mathcal{V}_k$ with $|\mathcal{V}_k|\geq2$, we gather its view features from the shared representation along the view dimension:
\begin{equation}
    \mathbf{Z}_k=\mathbf{Z}[\mathcal{V}_k,:,:]
    \in\mathbb{R}^{|\mathcal{V}_k|\times P\times D}.
    \label{eq:method-group-selection}
\end{equation}
All patch tokens of each selected view are retained, and the decoder receives only the features of that group.
Cross-view attention during this decoding stage therefore operates within the selected group.
We denote this decoding stage and its camera, point, and confidence predictors collectively by $\mathcal{D}_{\omega}$.
The same decoder is applied separately to each group:
\begin{equation}
    \mathcal{R}_k
    =\mathcal{D}_{\omega}(\mathbf{Z}_k)
    =\{(\mathbf{T}_i^{(k)},\mathbf{P}_i^{(k)},\mathbf{C}_i^{(k)})\}_{i\in\mathcal{V}_k},
    \label{eq:method-group-decoding}
\end{equation}
where $\mathbf{T}_i^{(k)}$ is a camera pose, $\mathbf{P}_i^{(k)}$ is a dense point map, and $\mathbf{C}_i^{(k)}$ is its confidence map.
Outputs are associated with the original view indices used to gather $\mathbf{Z}_k$.
Disjoint groups are reconstructed separately; any alignment between their reconstructions belongs to the downstream mapping stage.

The shared representation $\mathbf{Z}$ is computed once and retained throughout graph prediction and reconstruction.
Each group reuses a subset of these jointly aggregated features, so group decoding does not require rerunning $\mathcal{B}_{\theta}$ on the original images.
The computation therefore consists of one shared backbone pass, one verification-head evaluation, and decoding for each non-singleton group.

\subsection{Training Objectives and Strategy}
\label{sec:method-training}

We train the verification head with two complementary objectives: a classification loss that supervises direct overlap between image pairs, and a contrastive loss that regularizes the shared view descriptors $\mathbf{v}_i$ using scene identity.

\paragraph{Classification loss.}
For each supervised pair $(i,j)\in\Omega$ within the input collection, let $t_{ij}\in\{0,1\}$ indicate whether the two views have direct geometric overlap.
Given the predicted covisibility probability $S_{ij}$ from Eq.~\ref{eq:method-pair-scores}, we minimize binary cross-entropy:
\begin{equation}
    \mathcal{L}_{\mathrm{cls}}
    =-\frac{1}{|\Omega|}\sum_{(i,j)\in\Omega}
    \left[t_{ij}\log S_{ij}+(1-t_{ij})\log(1-S_{ij})\right].
    \label{eq:method-classification-loss}
\end{equation}
This loss supervises pairwise overlap rather than scene indentity: two views from the same scene can still form a negative pair if they do not overlap directly.

\paragraph{Contrastive loss.}
To complement pairwise overlap supervision, we apply a supervised contrastive loss~\citep{khosla2020supcon} to the same pooled descriptors $\mathbf{v}_i$ used by the pairwise prediction head, which is similar to visual place recongnition~\citep{Berton_2025_MegaLoc} methods.
Positive samples are defined by ground truth scene identity $y_i$ under each dataset's labeling strategy, rather than by direct overlap.
For anchor $i$, we define $\mathcal{P}(i)=\{j\in\mathcal{V}\setminus\{i\}:y_j=y_i\}$ and retain anchors with at least one positive partner in $\mathcal{A}=\{i\in\mathcal{V}:|\mathcal{P}(i)|>0\}$.
Using normalized descriptors $\widehat{\mathbf{v}}_i=\mathbf{v}_i/\|\mathbf{v}_i\|_2$, the loss is
\begin{equation}
    \mathcal{L}_{\mathrm{con}}
    =-\frac{1}{N}\sum_{i\in\mathcal{A}}
    \frac{1}{|\mathcal{P}(i)|}\sum_{p\in\mathcal{P}(i)}
    \log\frac{\exp(\widehat{\mathbf{v}}_i^{\top}\widehat{\mathbf{v}}_p/\tau)}
    {\sum_{a\in\mathcal{V}\setminus\{i\}}\exp(\widehat{\mathbf{v}}_i^{\top}\widehat{\mathbf{v}}_a/\tau)},
    \label{eq:method-contrastive-loss}
\end{equation}
where $\tau$ is the temperature.
The loss encourages descriptors from the same scene to align relative to other views in the collection. The definition of $\mathcal{P}$ for each dataset is provided in Appendix~\ref{app:datasets}.

\paragraph{Training strategy.}
We combine the two losses as
\begin{equation}
    \mathcal{L}=\mathcal{L}_{\mathrm{cls}}+\lambda\mathcal{L}_{\mathrm{con}},
    \label{eq:method-total-loss}
\end{equation}
where $\lambda$ controls the strength of contrastive regularization.
We conduct exhaustive ablations studies as shown in Table~\ref{tab:ablation} to estimate the optimal $\lambda$. We also explore the partial fine-tuning and full-layer LoRA~\citep{hu2021lora} strategies for the backbone $\mathcal{B}_{\theta}$.

%% file: figures/system_pipeline.tex
\begin{figure}[h]
    \centering
    \includegraphics[width=0.98\linewidth]{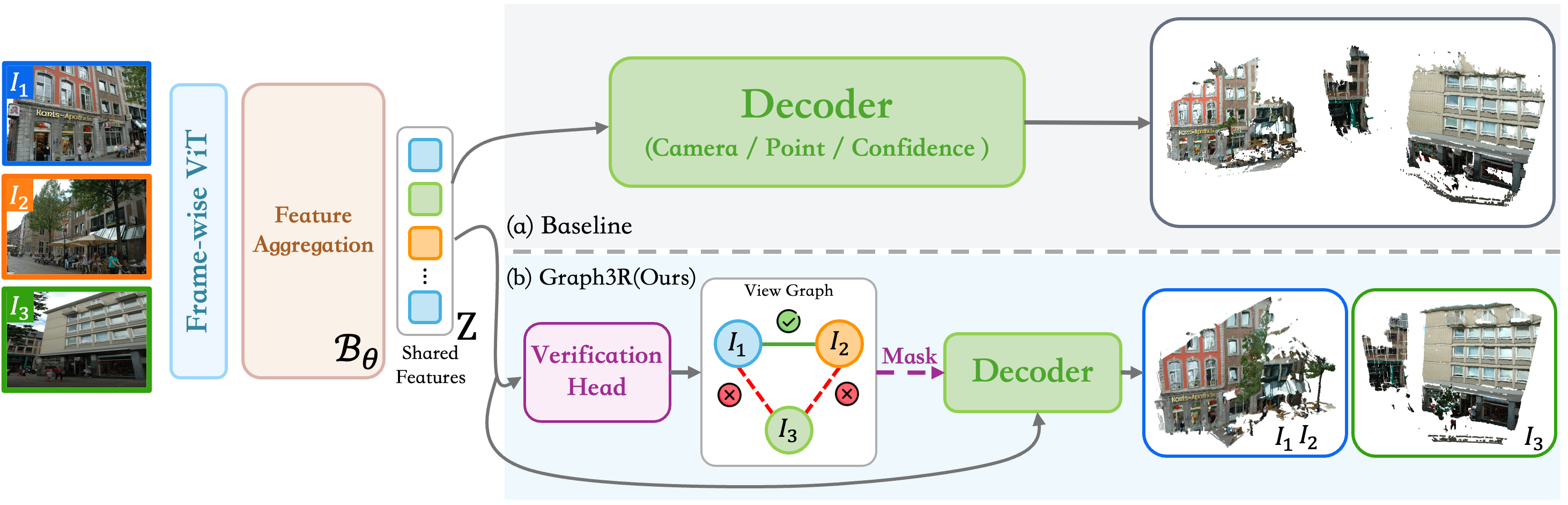}
    \caption{
    \textbf{System overview of \algname.}
    (a) The $\pi^3$ baseline~\citep{wang2025pi3}.
    (b) \algname introduces a verification head that predicts
    pairwise covisibility before geometry decoding.
    These predictions are used to construct
    a view graph, which guides the graph-conditioned reconstruction as described in Sec.~\ref{sec:method_graph-conditioned}, enabling each scene to be reconstructed individually.
}
    \label{fig:method}
\end{figure}

%% file: figures/verification_head.tex
\begin{figure}
    \vspace{-1em}
    \centering
    \includegraphics[width=0.8\linewidth]{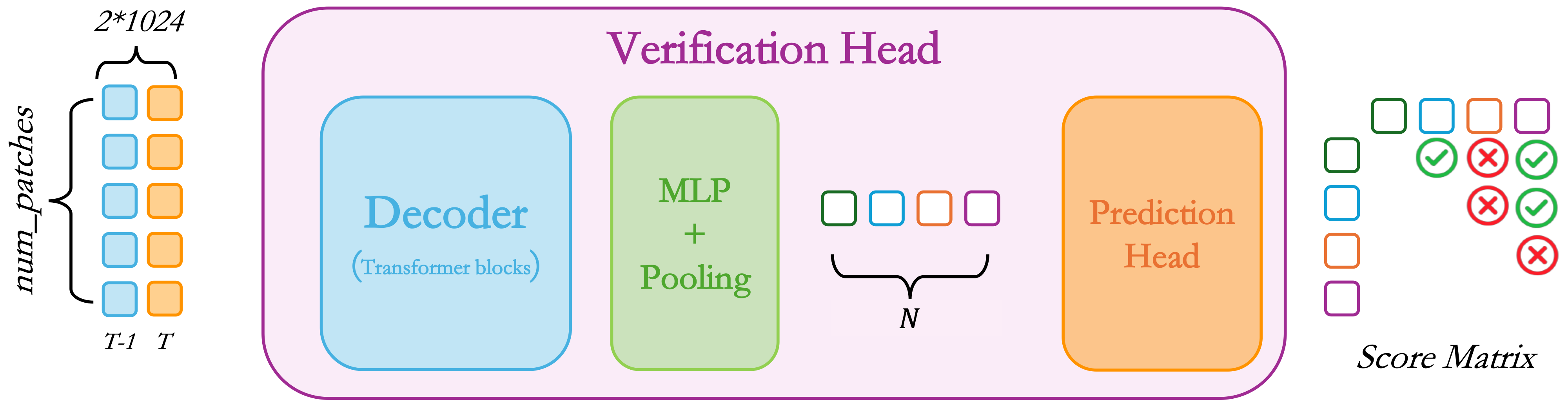}
    \caption{\textbf{Architecture of the verification head.} The head takes the shared feature representation $\mathbf{Z}$ as input and predicts pairwise covisibility scores $\mathbf{S}$. The scores are used to construct a view graph $\mathcal{G}$, which defines the mask of each scene group. The masks are applied to the shared representation $\mathbf{Z}$ prior to the geometry decoder, enabling graph-conditioned reconstruction for each scene group.
    }
    \label{fig:verification-head-architecture}
\end{figure}

%% file: sections/exp.tex
\section{Experiments} \label{experiments}

We evaluate \algname through four complementary experiments:
(a) controlled feature reuse to assess the effect of shared features on
graph-conditioned reconstruction;
(b) view graph initialization to measure association accuracy;
(c) mixed-scene reconstruction to evaluate end-to-end geometric prediction;
and (d) real-world multi-session mapping to assess cross-session localization
and computational efficiency.

\paragraph{Datasets.}
We use GSV-Cities~\citep{ali2022gsv} and 360Loc~\citep{huang2024360loc} to
train the verification head and evaluate view graph initialization.
ETH3D~\citep{schops2017multi} provides high-fidelity ground-truth depth for
controlled feature-reuse evaluation and supports multi-view reconstruction
benchmarks under Easy and Hard overlap splits.
OpenNavMap~\citep{jiao2026opennavmap} provides crowdsourced visual mapping
data from two geographically distinct regions, with up to 55 sessions per
region, for evaluating pose recovery and cross-session localization.
The selection of these datasets and split details are provided in Appendix~\ref{app:datasets}.

\paragraph{Baselines.}
We use $\pi^3$~\citep{wang2025pi3} and VGGT~\citep{wang2025vggt} as
standard feed-forward reconstruction baselines. Neither explicitly verifies
view associations, limiting their applicability to uncurated image collections.
We also compare with GLUEMAP~\citep{pan2026gluemap}, a closely related system
that constructs a view graph using Doppelgangers++~\citep{xiangli2025doppelgangers}
and uses it to guide $\pi^3$ reconstruction.
We denote its view graph initialization component by GLUEMAP$^\dagger$.
For the reconstruction and pose estimation benchmarks, we omit GLUEMAP's
post-optimization module for fair comparison with the direct feed-forward outputs.

\paragraph{Implementation details.}
We construct view graphs by retaining edges with $S_{ij}>\tau_s$ and
partitioning the resulting graph into connected components, with
$\tau_s=0.5$ across all benchmarks.
Following the ablation study in Sec.~\ref{exp:ablation}, we use full-layer
LoRA, a GSV-Cities and 360Loc training mixture, and a contrastive loss weight
of $\lambda=0.25$. The training is conducted on 16 NVIDIA H200 GPUs, and evaluations are conducted on a single NVIDIA A800 GPU.

\subsection{Controlled Feature Reuse}
\algname aims to unify view graph initialization and feed-forward
reconstruction through a shared feature representation.
However, as shown
in Fig.~\ref{fig:method}, views from unrelated scenes are processed together
by $\mathcal{B}_{\theta}$, whose global attention aggregates across all input
views.
This aggregation may introduce cross-scene noise into the
$\mathbf{Z}$ passed to the decoders, even when each scene is decoded
separately with masks.
In principle, jointly training the backbone and geometry decoder
with the verification head could help the model learn representations
that are robust to such interference.
This would require substantial training data and computational resources, which are beyond the scope of the question that \algname aims to answer.
We therefore conduct a controlled experiment to
assess how shared features affect graph-conditioned reconstruction.

We compare \textit{Individual} and \algname on
ETH3D~\citep{schops2017multi}, using identical $\pi^3$ weights and
ground-truth scene groups to isolate the effect of feature sharing.
\textit{Individual} feeds each scene to the backbone and decoder
independently, so no information is mixed across scenes in
$\mathcal{B}_{\theta}$. In contrast, \algname processes the entire
mixed-scene collection in a single backbone pass, then decodes each scene separately.
We vary the number of scenes $K$ from 2 to 4, more details
are provided in Appendix~\ref{app:feature-reuse}.
As shown in Table~\ref{tab:feature-reuse}, the degradation remains modest
as $K$ increases.
This suggests that shared features retain sufficient geometric information
for separate scene reconstruction despite cross scene interactions in
the $\mathcal{B}_{\theta}$. Feature reuse also avoids a separate backbone pass for
each scene, contributing to the efficiency of \algname as reported in Table~\ref{tab:runtime}, which can achieve nearly 30\% speedup.
\input{tables/exp_feature_reuse.tex}

\subsection{View Graph Initialization}
As discussed in GLUEMAP~\citep{pan2026gluemap}, accurate view graph initialization is key to accurate and robust reconstruction.
Following Doppelgangers++~\citep{xiangli2025doppelgangers} and GV-Bench~\citep{yu2024gv}, we evaluate view graph initialization as a pairwise classification task, using average precision (AP) and maximum recall at 100\% precision (MR) as the evaluation metrics.
Table~\ref{tab:view-graph-initialization} summarizes the results, with more detailed benchmarks provided in Sec.~\ref{app:multi-view-verification}.
\algname achieves the best average AP and MR across the benchmarks.
To achieve optimal performance with limited training resources, we conduct an ablation study to estimate the best training setup.
\input{tables/exp_view_graph.tex}

\paragraph{Ablation Study.}\label{exp:ablation}
Table~\ref{tab:ablation} studies the training data mixture, contrastive loss weight, and trainable modules of the backbone aggregator $\mathcal{B}_{\theta}$.
Mixing GSV-Cities and 360Loc improves generalization. With this mixture, contrastive regularization substantially increases MR, and full-layer LoRA achieves the best accuracy with fewer trainable parameters than partial finetuning. We therefore use the 4:1 data mixture, $\lambda=0.25$, and full-layer LoRA in subsequent experiments.

\input{tables/exp_ablation_horizontal.tex}
\subsection{Reconstruction}
We assess how predicted view graph affects reconstruction quality in two aspects: multi-view depth consistency and pose estimation accuracy for uncurated images.
\paragraph{Multi-view depth consistency.}
We use MVS, AbsRel, and $\delta_1$ to measure the consistency of predicted depths across views. With reference to the reprojection-based consistency checks in MVSNet~\citep{yao2018mvsnet} and the multi-view consistency evaluation in GeoVideo~\citep{bai2025geovideo}, we define MVS as the reprojection-based cross-view depth-consistency score used in this experiment. Its detailed definition and aggregation protocol are provided in Appendix~\ref{app:eth3d-verification-consistency}.
We conduct evaluation on ETH3D~\citep{schops2017multi}, using $\pi^3$ as the feed-forward reconstruction method for all methods for fair comparison. Table~\ref{tab:exp_depth} shows that \algname achieves the best performance across all metrics in both \textit{Easy} and \textit{Hard} splits, improving average MVS from 0.182 to 0.223 over RobustPi3 while also reducing AbsRel and increasing $\delta_1$. More detailed results including the benchmark construction are provided in Appendix~\ref{app:eth3d-verification-consistency}.
\input{tables/exp_depth.tex}

\paragraph{Pose estimation accuracy.}
We evaluate the accuracy in Table~\ref{tab:pose_estimation} that requires each method to reject unrelated views and recover accurate poses from related ones. \algname achieves the best performance on ETH3D and under the $20^\circ$/5\,m criterion on 360Loc, while remaining competitive with GLUEMAP on OpenNavMap. Appendix~\ref{app:pose-estimation} provides a more comprehensive discussion.
\input{tables/exp_pose.tex}

\subsection{Multi-session Mapping}
Multi-session mapping requires reliable cross-session associations to accurately align and merge maps. Following the OpenNavMap, we assess this capability through candidate reranking (Table~\ref{tab:opennavmap_rerank}) and pose estimation (Table~\ref{tab:pose_estimation}), and report computational efficiency in Table~\ref{tab:runtime} and Fig.~\ref{fig:efficiency}. For reranking, we first retrieve 8 candidates per query using CosPlace~\citep{berton2022rethinking}. \algname achieves the best reranking performance in both regions.
In Table~\ref{tab:runtime}, s/batch denotes runtime in seconds per batch of 9 images (1 query and 8 retrieved candidates). \textit{All groups} decodes every predicted connected component, whereas \textit{Query group} decodes only the component containing the query image.
\algname is the fastest method in both settings, with approximately \tb{27\%} and \tb{20\%} lower latency, respectively, than the next-fastest method. In addition, \algname achieves competitive performance with GLUEMAP in pose estimation on OpenNavMap, with runtime speedups of approximately $\tb{29.3}\times$ and $\tb{35.9}\times$ over GLUEMAP when decoding all predicted groups and only the query group, respectively.
Although its peak GPU memory usage is lower than that of RobustVGGT, it remains considerably higher than that of the other baselines.
Full benchmark statistics and details are provided in Appendix~\ref{app:opennavmap-runtime}.
Additional multi-session mapping visualizations are provided in Appendix~\ref{app:multi-session-visualization}.

\input{tables/exp_opennavmap_rerank.tex}
\input{figures/effiency.tex}
\input{tables/exp_opennavmap_runtime.tex}

%% file: tables/exp_feature_reuse.tex
\begin{table}[!htbp]
  \centering
  \small
    \caption{
      \textbf{Controlled feature reuse on ETH3D.}
      We separately report AbsRel$\downarrow$ and $\delta_1$(\%)$\uparrow$
      for independent decoding and group decoding from one shared backbone
      representation.
      $\Delta$ reports relative percentage changes in each metric in comparison to the \textit{Individual} baseline.
    }
    \label{tab:feature-reuse}
    \renewcommand{\arraystretch}{0.70}
    \setlength{\tabcolsep}{17pt}
    \begin{tabular}{@{}lcccccc@{}}
      \toprule
      \multirow{2}{*}{$K$}
      & \multicolumn{2}{c}{Individual}
      & \multicolumn{2}{c}{\algname}
      & \multicolumn{2}{c}{$\Delta$ (\%)} \\
      \cmidrule(lr){2-3}\cmidrule(lr){4-5}\cmidrule(l){6-7}
      & AbsRel$\downarrow$ & $\delta_1$(\%)$\uparrow$
      & AbsRel$\downarrow$ & $\delta_1$(\%)$\uparrow$
      & AbsRel$\downarrow$ & $\delta_1$(\%)$\uparrow$ \\
      \midrule
      2 & 0.098 & 92.9 & 0.098 & 91.9 & 0.00 & $-1.08$ \\
      3 & 0.097 & 93.4 & 0.097 & 92.1 & 0.00 & $-1.39$ \\
      4 & 0.098 & 93.2 & 0.102 & 91.4 & $+4.08$ & $-1.93$ \\
      \bottomrule
    \end{tabular}
    \vspace{-1em}
\end{table}

%% file: tables/exp_view_graph.tex
\begin{table}[h]
	\centering
	\caption{
	\textbf{Benchmark of View Graph Initialization.}
	}
	\label{tab:view-graph-initialization}
	\resizebox{0.95\linewidth}{!}{
	\begin{tabular}{@{}lcccccccc@{}}
		\toprule
		\multirow{2}{*}{Method}
		& \multicolumn{2}{c}{GSV-Cities}
		& \multicolumn{2}{c}{360Loc}
		& \multicolumn{2}{c}{ETH3D}
		& \multicolumn{2}{c}{Average} \\
		\cmidrule(lr){2-3}\cmidrule(lr){4-5}\cmidrule(lr){6-7}\cmidrule(lr){8-9}
		& mAP $\uparrow$ & MR $\uparrow$ & mAP $\uparrow$ & MR $\uparrow$ & mAP $\uparrow$ & MR $\uparrow$ & mAP $\uparrow$ & MR $\uparrow$ \\
		\midrule
		GLUEMAP$^\dagger$~\citep{pan2026gluemap} & \underline{91.17} & \underline{27.29} & \underline{96.13} & \underline{74.79} & 96.86 & \tb{93.17} & \underline{94.72} & \underline{65.08} \\
		RobustVGGT~\citep{han2025emergent} & 70.85 & 3.46 & 91.89 & 15.36 & 86.58 & 49.97 & 83.11 & 22.93 \\
		RobustPi3~\citep{han2025emergent} & 74.89 & 0.20 & 93.46 & 6.31 & \underline{97.76} & 65.38 & 88.70 & 23.96 \\
		\tb{\algname (Ours)} & \tb{98.98} & \tb{47.17} & \tb{99.65} & \tb{83.64} & \tb{99.38} & \underline{91.41} & \tb{99.34} & \tb{74.07} \\
		\bottomrule
	\end{tabular}
	}
\end{table}

%% file: tables/exp_ablation_horizontal.tex
\begin{table}[t]
  \centering
  \caption{\textbf{Ablations of the verification head.}
    All the \textit{Data Mixture} uses
    $\lambda=0$, \textit{Contrastive Loss Weight} and \textit{Trainable Modules} use the 4:1
    mixture, with LoRA and $\lambda=0.25$ fixed, respectively.
    Params. counts include the verification head.}
  \label{tab:ablation}
  \setlength{\tabcolsep}{3pt}
  \resizebox{0.95\linewidth}{!}{
  \renewcommand{\arraystretch}{0.85}
  \begin{tabular*}{\linewidth}{@{\extracolsep{\fill}}lccc|lcc|lccc@{}}
    \toprule
    \multicolumn{4}{c|}{Data Mixture}
      & \multicolumn{3}{c|}{Contr. Loss Weight}
      & \multicolumn{4}{c}{Trainable Modules} \\
    \cmidrule(lr){1-4}\cmidrule(lr){5-7}\cmidrule(lr){8-11}
    Backbone & Data & AP$\uparrow$ & MR$\uparrow$
      & $\lambda$ & AP$\uparrow$ & MR$\uparrow$
      & Backbone & Params. & AP$\uparrow$ & MR$\uparrow$ \\
    \midrule
    Frozen & GSV & 93.75 & 37.32
      & 0 & 99.14 & 57.25
      & Frozen & 69M & 98.32 & 57.07 \\
    Frozen & 4:1 & 98.11 & \textbf{57.46}
      & 0.1 & 99.22 & 64.29
     &   \cornerarrow FT2 & 103M & 97.97 & 53.44 \\
    LoRA & GSV & 97.67 & 53.07
      & 0.25 & \textbf{99.31} & \textbf{73.06}
      &  \cornerarrow FT4 & 128M & 98.70 & 38.49 \\
    LoRA & 4:1 & \textbf{99.14} & 57.25
      & 0.5 & 99.09 & 54.82
      &  \cornerarrow LoRA & 80M & \textbf{99.31} & \textbf{73.06} \\
    \bottomrule
  \end{tabular*}
  }
\end{table}

%% file: tables/exp_depth.tex
\begin{table}[!htbp]
\centering
\caption{
\tb{Multi-view depth consistency} on ETH3D~\citep{schops2017multi}.
}
	\label{tab:exp_depth}
	\setlength{\tabcolsep}{3pt}
  \renewcommand{\arraystretch}{0.95}
	\resizebox{0.95\linewidth}{!}{
	\begin{tabular}{@{}lccccccccc@{}}
		\toprule
		\multirow{2}{*}{Method} & \multicolumn{3}{c}{\textit{Easy}} & \multicolumn{3}{c}{\textit{Hard}} & \multicolumn{3}{c}{Average} \\
		\cmidrule(lr){2-4}\cmidrule(lr){5-7}\cmidrule(lr){8-10}
		& MVS$\uparrow$ & AbsRel$\downarrow$ & $\delta_1$(\%)$\uparrow$
		& MVS$\uparrow$ & AbsRel$\downarrow$ & $\delta_1$(\%)$\uparrow$
		& MVS$\uparrow$ & AbsRel$\downarrow$ & $\delta_1$(\%)$\uparrow$ \\
		\midrule
		RobustPi3~\citep{han2025emergent} & \underline{0.230} & \underline{0.287} & \underline{73.6} & \underline{0.133} & \underline{0.394} & \underline{65.9} & \underline{0.1815} & \underline{0.3405} & \underline{69.75} \\
		GLUEMAP~\citep{pan2026gluemap} & 0.127 & 0.549 & 61.2 & 0.078 & 0.586 & 59.9 & 0.1025 & 0.5675 & 60.55 \\
		\textbf{\algname{} (ours)} & \textbf{0.271} & \textbf{0.264} & \textbf{77.5} & \textbf{0.175} & \textbf{0.321} & \textbf{70.9} & \textbf{0.2230} & \textbf{0.2925} & \textbf{74.20} \\
		\bottomrule
	\end{tabular}
	}
\end{table}

%% file: tables/exp_pose.tex
\begin{table}[!htbp]
  \centering
  \small
  \caption{
    \tb{Benchmark of pose estimation accuracy.} We report the accuracy at different thresholds.
  }
  \label{tab:pose_estimation}
  \setlength{\tabcolsep}{6.5pt}
  \renewcommand{\arraystretch}{0.8}
  \begin{tabular}{lcccccc}
    \toprule
    \multirow{2}{*}{Method} & \multicolumn{2}{c}{ETH3D} & \multicolumn{2}{c}{OpenNavMap} & \multicolumn{2}{c}{360Loc} \\
    \cmidrule(lr){2-3}\cmidrule(lr){4-5}\cmidrule(lr){6-7}
    & $10^\circ$/1m  & $20^\circ$/5m     
    & $10^\circ$/1m  & $20^\circ$/5m 
    & $10^\circ$/1m  & $20^\circ$/5m \\
    \midrule
    Pi3~\citep{wang2025pi3}
    & 27.55 & 41.04 & 19.87 & 31.17 & 1.86 & 16.86
    \\
    VGGT~\citep{wang2025vggt}
    & 8.43 & 22.99 & 8.89 & 28.60 & 0.25 & 12.06
    \\
    \midrule
    RobustPi3~\citep{han2025emergent}
    & \underline{73.08} & \underline{85.71} & 60.92 & 68.36 & \textbf{54.07} & 64.71
    \\
    RobustVGGT~\citep{han2025emergent}
    & 54.90 & 62.82 & 48.13 & 70.63 & 49.75 & 62.75
    \\
    GLUEMAP~\citep{pan2026gluemap}
    & 60.00 & 72.04 & \textbf{67.83} & \textbf{79.72} & 52.79 & \underline{66.37}
    \\
    \tb{Graph3R(Ours)}
    & \textbf{76.19} & \textbf{88.09} & \underline{67.65} & \underline{75.86} & \underline{53.82} & \textbf{72.70}
    \\
    \bottomrule
  \end{tabular}
\end{table}

%% file: tables/exp_opennavmap_rerank.tex
\begin{table}[!htbp]
  \centering
  \small
  \caption{\tb{Cross-session localization on OpenNavMap.}
  All methods rerank on CosPlace retrieved candidates.}
  \label{tab:opennavmap_rerank}
  \setlength{\tabcolsep}{6pt}
  \renewcommand{\arraystretch}{0.85}
  \begin{tabular}{@{}l*{8}{c}@{}}
    \toprule
     \multirow{2}{*}{Method} & \multicolumn{4}{c}{\textit{Region A}} & \multicolumn{4}{c}{\textit{Region B}} \\
    \cmidrule(lr){2-5}\cmidrule(lr){6-9}
    & R@1 & R@3 & R@5 & R@8
            & R@1 & R@3 & R@5 & R@8\\
    \midrule
    CosPlace~\citep{berton2022rethinking}
      & 63.93 & 69.48 & 73.01 & 76.04
      & 66.05 & 71.60 & 73.95 & 76.49 \\
    RobustPi3~\citep{han2025emergent}
      & 65.83 & 73.77 & 75.28 & --
      & 70.35 & \underline{74.30} & \underline{75.56} & -- \\
    RobustVGGT~\citep{han2025emergent}
      & 71.25 & 74.65 & 75.28 & --
      & \underline{71.35} & 74.21 & 75.26 & -- \\
    GLUEMAP$^{\dagger}$~\citep{pan2026gluemap}
      & \underline{72.26} & \underline{75.41} & \underline{75.91} & --
      & 71.33 & 73.74 & 75.19 & -- \\
    \textit{\tb{Graph3R(Ours)}}
      & \textbf{74.15} & \textbf{75.54} & \textbf{76.04} & --
      & \textbf{72.88} & \textbf{75.00} & \textbf{75.79} & -- \\
    \bottomrule
  \end{tabular}
\end{table}

%% file: figures/effiency.tex
\begin{figure}[!htbp]
\centering
\includegraphics[width=0.85\linewidth]{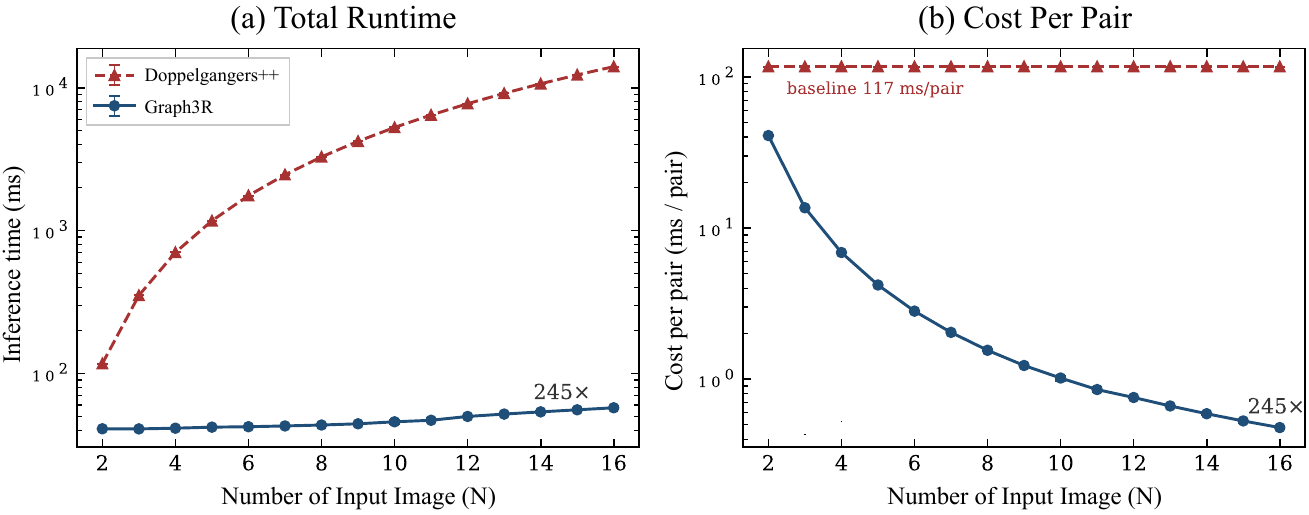}
\caption{\textbf{Efficiency comparison} with Doppelgangers++~\citep{xiangli2025doppelgangers}. Our method achieves approximately 245 $\times$speedup as the number of input images $N=16$.
}
\label{fig:efficiency}
\end{figure}

%% file: tables/exp_opennavmap_runtime.tex
\begin{table}[H]
  \centering
  \caption{
    \textbf{Computational efficiency on OpenNavMap.}
    End-to-end latency and peak GPU memory (GB)
    for $N=9$ images per batch, decoding all predicted groups or only the group containing the query.
  }
  \label{tab:runtime}
  \renewcommand{\arraystretch}{0.85}
  \resizebox{0.93\linewidth}{!}{
  \begin{tabular}{@{}lcccc@{}}
    \toprule
    \multirow{2}{*}{Method}
      & \multicolumn{2}{c}{All groups}
      & \multicolumn{2}{c}{Query group} \\
    \cmidrule(lr){2-3}\cmidrule(lr){4-5}
      & s/batch $\downarrow$ & Peak Memory $\downarrow$
      & s/batch $\downarrow$ & Peak Memory $\downarrow$ \\
    \midrule
    RobustPi3~\citep{han2025emergent} & \underline{0.218} & \underline{4.727} & \underline{0.158} & \underline{4.727} \\
    RobustVGGT~\citep{han2025emergent} & 0.693 & 11.065 & 0.421 & 11.065 \\
    GLUEMAP~\citep{pan2026gluemap} & 4.663 & \textbf{4.635} & 4.564 & \textbf{4.637} \\
    \algname (Ours) & \textbf{0.159} & 8.087 & \textbf{0.127} & 8.087 \\
    \bottomrule
  \end{tabular}%
  }
\end{table}

%% file: sections/conclusion.tex
\section{Conclusion}
\label{conclusion}

We presented \algname, which unifies view graph initialization and feed-forward reconstruction through shared geometric features to reconstruct image collections containing disjoint scenes.
A learned verification head predicts pairwise covisibility, and the resulting graph guides separate pose and geometry decoding for each scene group using features from a single backbone pass.
Controlled experiments show that feature reuse incurs only a modest loss in depth accuracy, while verification and reconstruction benchmarks demonstrate the benefits of learned grouping for uncurated inputs.
On OpenNavMap, \algname improves cross-session localization and achieves competitive pose estimation accuracy with GLUEMAP, with approximately $35.9\times$ runtime speedup when decoding only the query group.
It achieves the lowest latency among the compared methods under the evaluated configurations, supporting its use as a front end for multi-session mapping.
Grouping errors can still affect reconstruction by merging unrelated views or separating connected ones.
Future work will explore grouping strategies that account for prediction uncertainty and more diverse training data to improve generalization across scenes and capture conditions.
With sufficient computational resources, we also plan to jointly train the backbone and geometry decoder alongside the verification head on mixed-scene image collections, aiming to preserve reconstruction accuracy despite feature mixing across unrelated scenes.

%% file: appendix.tex
\clearpage
\appendix
\section{Supplementary Material}

\input{sections/appendix_datasets.tex}

\subsection{Comparison of Different SfM Paradigms}
\label{app:paradigm-comparison}
Table~\ref{tab:related-works} complements Fig.~\ref{fig:paradigms} with the
association features, decision rules, and reconstruction stages of representative
methods. Feature context denotes the images used to compute association
features, not the input requirements of the reconstruction backbone.
\input{tables/related_works.tex}

\subsection{Controlled Feature Reuse}
\label{app:feature-reuse}
Table~\ref{tab:feature-reuse} uses the ETH3D Easy (clique) setting with
$K\in\{2,3,4\}$ scenes, four views per scene, and 20 fixed episodes per setting.
Both variants use oracle scene groups and identical $\pi^3$ weights at
$224\times224$ resolution. Independent reconstruction computes features
separately for each scene; shared reconstruction gathers each group's features
from a single mixed-collection backbone pass. Depth is evaluated against
ground truth with per-scene median-ratio scale alignment, a shared GT mask,
and equal episode weighting. This GT-depth evaluation is distinct from the
cross-view consistency evaluation in Appendix~\ref{app:eth3d-verification-consistency}.

\subsection{View Graph Initialization}
\label{app:multi-view-verification}
Table~\ref{tab:multi-view-verification-full} expands
Table~\ref{tab:view-graph-initialization} with all four GSV-Cities validation
splits and the attention-only ($\mathcal{A}$) and feature-only ($\mathcal{F}$)
robust baselines. Its Average weights the four GSV-Cities columns and 360Loc
equally; it is not the three-dataset average in the main table, which uses
GSV-Cities Full, 360Loc, and ETH3D. The ETH3D breakdown is provided in
Table~\ref{tab:eth3d-verification-consistency}.
\input{tables/exp_multi_full.tex}

\paragraph{Ablation Study.}
Table~\ref{tab:ablation-full} gives per-split results for the configurations
in Table~\ref{tab:ablation}. The data-composition comparison uses
classification loss alone ($\lambda=0$); the contrastive-weight comparison
uses full-layer LoRA and the 4:1 mixture; the adaptation comparison uses
the 4:1 mixture and $\lambda=0.25$. FT2 and FT4 update the last two and
four backbone layers, respectively. The reported checkpoints have different
step counts, so these results compare the saved configurations rather than
a matched-step training budget.
\input{tables/appendix_ablation.tex}
\FloatBarrier

\subsection{Multi-view Depth Consistency}
\label{app:eth3d-verification-consistency}
Table~\ref{tab:eth3d-depth-consistency-average-full} extends
Table~\ref{tab:exp_depth} with RobustVGGT as an additional grouping scorer.
Table~\ref{tab:eth3d-verification-consistency} provides the per-$K$ breakdown,
including raw-score AP/MR and thresholded edge F1. All scorers share
$\pi^3$ reconstruction weights and features; only the predicted grouping
changes. Easy requires pairwise within-scene overlap, whereas Hard requires
only a connected within-scene overlap graph.

\paragraph{MVS definition.}
We evaluate cross-view geometric consistency using a reprojection-based
score, denoted by MVS. For a source pixel $\mathbf{p}$ in view $i$, we
reproject its predicted depth into view $j$ using the predicted relative pose
$\mathbf{T}_j^{-1}\mathbf{T}_i$ and the camera intrinsics recovered from the
predicted point maps. Let
$\mathbf{u}_{i\rightarrow j}(\mathbf{p})$ and
$z_{i\rightarrow j}(\mathbf{p})$ denote the projected location and the
transformed depth in the target camera, respectively. We bilinearly sample
the target depth prediction as
\begin{equation}
  \widehat{D}_j(\mathbf{p})
  =
  D_j\!\left(\mathbf{u}_{i\rightarrow j}(\mathbf{p})\right).
  \label{eq:mvs-target-depth}
\end{equation}
The directional consistency score is
\begin{equation}
  c_{i\rightarrow j}
  =
  \frac{1}{H_iW_i}
  \sum_{\mathbf{p}\in\Omega_i}
  \mathbf{1}\!\left[
    v_{i\rightarrow j}(\mathbf{p})
    \land
    \left|
      \frac{
        z_{i\rightarrow j}(\mathbf{p})-\widehat{D}_j(\mathbf{p})
      }{
        \widehat{D}_j(\mathbf{p})
      }
    \right| < 0.05
  \right],
  \label{eq:mvs-directional}
\end{equation}
where $\Omega_i$ is the set of all $H_iW_i$ source pixels and
$v_{i\rightarrow j}(\mathbf{p})$ requires a positive source depth and a
projection inside the target image. Invalid projections and undefined depth
comparisons contribute zero. The normalization therefore uses all source
pixels, rather than only valid projections.

For an input episode $e$, let $\mathcal{E}_e$ contain all unordered view
pairs within the predicted connected components. We compute
\begin{equation}
  \mathrm{MVS}_e
  =
  \frac{1}{|\mathcal{E}_e|}
  \sum_{\{i,j\}\in\mathcal{E}_e}
  \frac{c_{i\rightarrow j}+c_{j\rightarrow i}}{2}.
  \label{eq:mvs-episode}
\end{equation}
We report the unweighted mean over episodes containing at least one retained
pair. Higher scores indicate greater cross-view consistency. MVS uses only
predicted geometry, without ground-truth depth or ground-truth scale
alignment. Because the retained pair set depends on the predicted grouping,
we interpret MVS together with the corresponding view-graph prediction
metrics.

The depth metrics measure agreement between predicted depths across views,
not accuracy against GT depth. AbsRel and $\delta_1$ use valid reprojections.
Retained pair sets differ between methods, so consistency should be read
alongside graph-prediction quality rather than as a coverage measure.
RobustVGGT attains the best Easy consistency and average MVS/$\delta_1$,
while \algname achieves the lowest average AbsRel.
\begin{table}[!ht]
\begin{minipage}{\linewidth}
\input{tables/exp_eth_recon_full.tex}
\end{minipage}
\end{table}
\input{tables/exp_eth3d_verification_consistency.tex}
\FloatBarrier

\subsection{Pose Estimation Accuracy}
\label{app:pose-estimation}
Table~\ref{tab:pose-system-aligned-full} restores the relative-pose metrics
omitted from Table~\ref{tab:pose_estimation}, under the same strict
score $>0.5$ verification protocol. AP-RPE is the median defined angular
error on accepted positive pairs. SA-RPE is the class-balanced system-aware
mean: correctly rejected negatives receive $0^\circ$, and decision or pose
failures receive $180^\circ$. Rotation and translation-direction errors
are reported separately. System accuracy is
$\tfrac12(\mathrm{TNR}+\mathrm{pose\ recall})$, with pose recall evaluated
at the joint rotation/translation thresholds shown in the table.
Thus, AP-RPE measures conditional pose quality, whereas SA-RPE and system
accuracy also account for verification failures.

OpenNavMap positives satisfy a GT camera-center distance of at most 25\,m
and relative orientation of at most $40^\circ$. Pi3, RobustPi3, and
GLUEMAP use Pi3 geometry, as does \algname; VGGT and
RobustVGGT use VGGT geometry.
\input{tables/appendix_pose.tex}
\FloatBarrier

\subsection{Multi-session Mapping}
\label{app:opennavmap-retrieval}
Table~\ref{tab:opennavmap-reranking-pose-full} expands
Table~\ref{tab:opennavmap_rerank} with the macro-average over Region A and Region B.
All methods rerank the same eight CosPlace candidates, so R@8 is a shared
retrieval ceiling, not a verifier-dependent result. Correctness uses
the same 25\,m/$40^\circ$ criterion as the OpenNavMap pose estimation evaluation.
These experiments evaluate cross-session localization for mapping;
they do not measure final map-merging or trajectory-alignment accuracy.
\input{tables/appendix_opennavmap.tex}

\clearpage
\subsection{Multi-session Mapping: Computational Efficiency}
\label{app:opennavmap-runtime}
Tables~\ref{tab:opennavmap-runtime-all-full}
and~\ref{tab:opennavmap-runtime-query-full} expand
Table~\ref{tab:runtime}. Each batch contains nine images and
36 unordered pairs. Methods use their own scores and the same strict
score $>0.5$ connected-component rule. Output scopes are all predicted
groups or only the group containing query image 0; predicted memberships
can differ across methods. Equivalent pair throughput is 36 times batch
throughput, with displayed values independently rounded.
Peak memory subtracts each run's measured idle baseline and uses decimal
GB ($10^9$ bytes); both average and maximum peaks are reported.

For GLUEMAP, Doppelgangers++ verification time and peak memory come from its original
36-pair scoring run; each group's Pi3 reconstruction was measured anew.
RobustPi3 and RobustVGGT verification and reconstruction were measured anew
for both scopes. \algname uses its validated dual-stream runs and cached
base geometry features.
\input{tables/appendix_runtime.tex}
\FloatBarrier

\clearpage
\subsection{View Graph Initialization: Pairwise Verification}
\label{app:pairwise-verification}
For a two-image comparison, Doppelgangers++ and \algname are trained on
210K pairs from VisymScenes~\citep{xiangli2025doppelgangers} and
Doppelgangers~\citep{cai2023doppelgangers}, separately from the multi-view
training setting in the main paper.
Table~\ref{tab:pairwise-verify} reports VisymScenes results and zero-shot
transfer to GV-Bench~\citep{yu2024gv};
Table~\ref{tab:pairwise-gvbench} gives the available per-condition breakdown.
\algname improves VisymScenes AP and MR over Doppelgangers++, but this
advantage does not carry over to GV-Bench, where MegaLoc has the best mean
AP and MR. These results distinguish in-domain pairwise verification from
cross-domain generalization.
\input{tables/appendix_pairwise.tex}
\FloatBarrier

\clearpage
\subsection{Visualization of Multi-session Mapping}
\label{app:multi-session-visualization}
\begin{figure}[htbp]
  \centering
  \includegraphics[width=\linewidth,height=0.80\textheight,keepaspectratio]{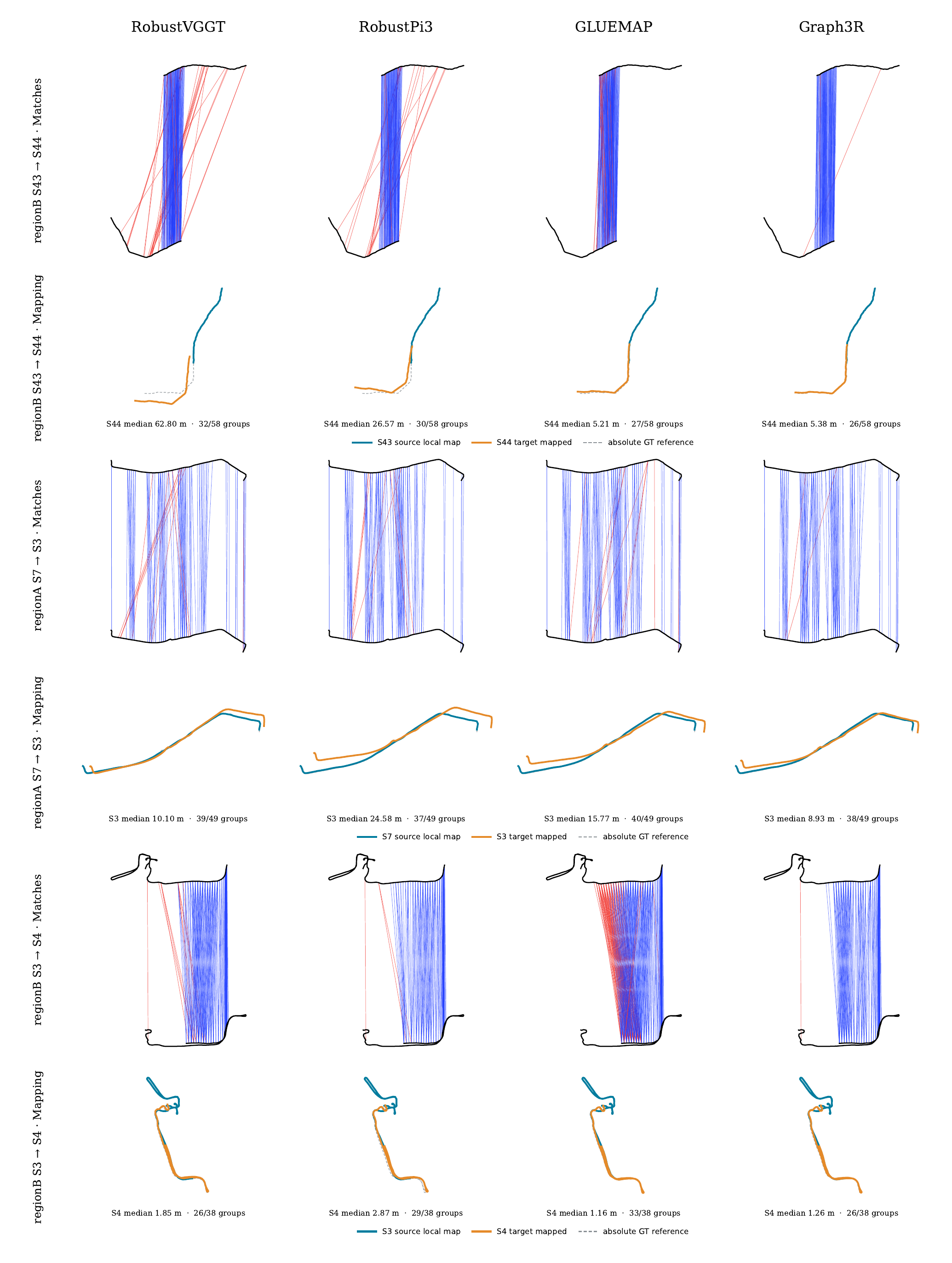}
  \caption{\textbf{Visualization of multi-session mapping on OpenNavMap.}
  Three cases are shown, each with cross-session associations above and
  the corresponding mapping results below.}
  \label{fig:multi-session-mapping-visualization}
\end{figure}

Fig.~\ref{fig:multi-session-mapping-visualization} visualizes multi-session
mapping results for three OpenNavMap cases, showing cross-session
associations and the corresponding mapping results for each method.

\FloatBarrier

%% file: sections/appendix_datasets.tex
\subsection{Datasets}
\label{app:datasets}
We distinguish the datasets used to train the multi-view verifier from
those used for geometric evaluation and the supplementary pairwise study.
The descriptions below specify their roles and the evaluation subsets
reported in the paper; task-specific metrics are defined in the corresponding
experimental sections.

\paragraph{GSV-Cities.}
GSV-Cities~\citep{ali2022gsv} provides place-labeled imagery for supervised
visual place recognition. We use it for multi-view verification training
and validation. Table~\ref{tab:view-graph-initialization} reports the Full
validation split, while Table~\ref{tab:multi-view-verification-full} additionally
reports Easy, Medium, and Hard. The selected training configuration mixes
GSV-Cities and 360Loc at a 4:1 ratio; the data-composition ablation compares
this mixture with GSV-Cities alone.

\paragraph{360Loc.}
360Loc~\citep{huang2024360loc} is an omnidirectional visual-localization
dataset with cross-device queries. It supplies the second source of
multi-view verification training data and a separate verification
evaluation reported alongside GSV-Cities. The pose experiment uses the
atrium subset, with results in Table~\ref{tab:pose-system-aligned-full}.
The verification and atrium pose evaluations are distinct tasks and should
not be interpreted as using an identical sample set.

\paragraph{ETH3D.}
ETH3D~\citep{schops2017multi} is used to evaluate controlled feature reuse,
view-graph prediction, depth consistency, and pose recovery.
For the controlled depth and graph experiments, each collection contains
$K\in\{2,3,4\}$ scenes with four views per scene, giving $4K$ input images.
Each setting uses 20 fixed episodes. Easy (clique) requires pairwise
within-scene overlap; Hard (connected) requires a connected within-scene
overlap graph without requiring all pairs to overlap.
Inputs are evaluated at $224\times224$ resolution.
Appendix~\ref{app:feature-reuse} describes the oracle-group GT-depth
comparison, and Appendix~\ref{app:eth3d-verification-consistency} describes
predicted-group cross-view consistency.

\paragraph{OpenNavMap.}
OpenNavMap~\citep{jiao2026opennavmap} supplies data from Region A and Region B
(Aria) for pose estimation evaluation, cross-session localization, and
runtime measurement. For pose estimation and cross-session localization, positive matches
require a ground-truth camera-center distance of at most 25\,m and relative
orientation of at most $40^\circ$. Cross-session localization reranks the
same eight CosPlace candidates for every method. Region-specific results
and their equal-weight macro-average are reported in
Table~\ref{tab:opennavmap-reranking-pose-full}.
The runtime protocol uses nine-image batches, as detailed in
Appendix~\ref{app:opennavmap-runtime}.

\paragraph{VisymScenes and Doppelgangers.}
These datasets are used for the supplementary two-image verification study,
not the GSV-Cities/360Loc multi-view training mixture.
Doppelgangers++ and our pairwise variants are trained on a combined 210K
pairs from VisymScenes~\citep{xiangli2025doppelgangers} and
Doppelgangers~\citep{cai2023doppelgangers}. VisymScenes provides the in-domain
evaluation in Table~\ref{tab:pairwise-verify}.

\paragraph{GV-Bench.}
GV-Bench~\citep{yu2024gv} is used for zero-shot evaluation of the pairwise
models trained above. Table~\ref{tab:pairwise-gvbench} reports Day, Night,
Weather, Season, Nordland, and Uacampus conditions, together with their
unweighted mean. This experiment tests transfer beyond the pairwise
training data; it is separate from the main multi-view verification
benchmark.

%% file: tables/related_works.tex
\begin{table}[htbp]
    \centering
    \caption{
        \textbf{View association and subsequent reconstruction.}
        DG denotes Doppelgangers; feature context refers to association.
    }
    \label{tab:related-works}
    \small
    \setlength{\tabcolsep}{3pt}
    \renewcommand{\arraystretch}{1.0}
    \begin{tabularx}{\linewidth}{@{}
        >{\raggedright\arraybackslash}p{0.23\linewidth}
        >{\raggedright\arraybackslash}p{0.16\linewidth}
        >{\raggedright\arraybackslash}p{0.23\linewidth}
        >{\raggedright\arraybackslash}X@{}}
        \toprule
        Method & Feature context & Association & Reconstruction \\
        \midrule
        MegaLoc & Per-image & Similarity & External stage \\
        RoMAv2 + RANSAC & Pairwise & Geometric verification & External stage \\
        DG / DG++ & Pairwise & Learned classifier & SfM \\
        \midrule
        RobustVGGT
        & Multi-view & Attention / features & Backbone rerun \\
        GLUEMAP & Image + pair & Retrieval + verification
        & Local inference + global optimization \\
        \midrule
        \rowcolor{gray!15}
        \algname (ours) & Multi-view & Learned classifier
        & Group decoding from retained features \\
        \bottomrule
    \end{tabularx}
\end{table}

%% file: tables/exp_multi_full.tex
\begin{center}
\captionof{table}{
\textbf{Full benchmark of view graph initialization.} We report mean average precision (mAP) and maximum recall at 100\% precision (MR) for every GSV-Cities validation split and 360Loc, together with the overall average.
}
\label{tab:multi-view-verification-full}
\setlength{\tabcolsep}{4pt}
\resizebox{0.95\linewidth}{!}{
\begin{tabular}{ll*{8}{c}|cc|cc}
	\toprule
	\multicolumn{2}{c}{\multirow{3}{*}{Method}}
	& \multicolumn{8}{c}{GSV-Cities}
	& \multicolumn{2}{c}{\multirow{2}{*}{360Loc}}
	& \multicolumn{2}{c}{\multirow{2}{*}{Average}} \\
	\cmidrule(lr){3-10}
	\multicolumn{2}{c}{}
	& \multicolumn{2}{c}{Easy}
	& \multicolumn{2}{c}{Medium}
	& \multicolumn{2}{c}{Hard}
	& \multicolumn{2}{c}{Full}
	& \multicolumn{2}{c}{}
	& \multicolumn{2}{c}{} \\
	\multicolumn{2}{c}{}
	& mAP $\uparrow$ & MR $\uparrow$
	& mAP $\uparrow$ & MR $\uparrow$
	& mAP $\uparrow$ & MR $\uparrow$
	& mAP $\uparrow$ & MR $\uparrow$
	& mAP $\uparrow$ & MR $\uparrow$
	& mAP $\uparrow$ & MR $\uparrow$ \\
	\midrule
	\multirow{2}{*}{\rotatebox[origin=c]{90}{\tiny Pairwise}}
	& RoMAv2~\citep{edstedt2025romav2} & 95.26 & 3.63 & 91.64 & 1.62 & 87.89 & 0.21 & 91.60 & 0.16 & 68.78 & 3.13 & 87.03 & 1.75 \\
	& GLUEMAP$^\dagger$~\citep{pan2026gluemap} & \tb{99.48} & 45.87 & 91.33 & 44.69 & 88.39 & 26.98 & 91.17 & 27.29 & 96.13 & 74.79 & 92.26 & 43.92 \\
	\midrule
	\multirow{7}{*}{\rotatebox[origin=c]{90}{\tiny Multi-view}}
	& RobustVGGT-$\mathcal{A}$ & 72.09 & 12.99 & 41.98 & 10.98 & 25.68 & 2.38 & 46.13 & 2.58 & 53.97 & 0.00 & 47.97 & 5.79 \\
	& RobustVGGT-$\mathcal{F}$ & 89.59 & 3.43 & 68.76 & 0.41 & 49.10 & 0.12 & 68.81 & 0.24 & 91.88 & 2.09 & 73.63 & 1.26 \\
	& RobustVGGT~\citep{han2025emergent} & 91.32 & 19.85 & 69.56 & 10.17 & 49.08 & 3.17 & 70.85 & 3.46 & 91.89 & 15.36 & 74.54 & 10.40 \\
	& RobustPi3-$\mathcal{A}$ & 84.71 & 0.88 & 63.05 & 1.11 & 43.11 & 0.04 & 60.32 & 0.03 & 82.59 & 6.09 & 66.76 & 1.63 \\
	& RobustPi3-$\mathcal{F}$ & 91.87 & 0.00 & 78.88 & 0.16 & 61.86 & 0.03 & 75.28 & 0.01 & 93.48 & 6.39 & 80.27 & 1.32 \\
	& RobustPi3~\citep{han2025emergent} & 93.34 & 0.10 & 78.67 & 0.52 & 60.46 & 0.35 & 74.89 & 0.20 & 93.46 & 6.31 & 80.16 & 1.50 \\
	& Ours & 99.43 & \tb{80.33} & \tb{99.18} & \tb{79.73} & \tb{98.25} & \tb{42.72} & \tb{98.98} & \tb{47.17} & \tb{99.65} & \tb{83.64} & \tb{99.10} & \tb{66.72} \\
	\bottomrule
\end{tabular}
}
\end{center}

%% file: tables/appendix_ablation.tex
\begin{table}[!ht]
\centering
\caption{\textbf{Per-split results for the main ablations.}
Each metric entry is AP/MR@100P (\%). Data is the GSV-Cities:360Loc ratio;
$\lambda=0$ disables the contrastive term.
The main-table mean weights the mean over four GSV-Cities splits and 360Loc equally.
Configurations shared by multiple ablations appear only once.}
\label{tab:ablation-full}
\setlength{\tabcolsep}{4pt}
\resizebox{\linewidth}{!}{
\begin{tabular}{@{}llrccccc@{}}
\toprule
\multirow{2}{*}{Adaptation} & \multirow{2}{*}{Data}
& \multirow{2}{*}{$\lambda$}
& \multicolumn{4}{c}{GSV-Cities} & \multirow{2}{*}{360Loc} \\
\cmidrule(lr){4-7}
& & & Easy & Medium & Hard & Full & \\
\midrule
Frozen & 1:0 & 0.0 & 99.45/67.62 & 99.55/58.38 & 98.38/39.16 & 98.59/19.58 & 88.51/28.46 \\
Frozen & 4:1 & 0.0 & 99.12/58.53 & 98.10/44.05 & 94.78/23.66 & 97.00/19.93 & 98.96/78.37 \\
LoRA & 1:0 & 0.0   & 99.83/84.86 & 99.69/90.41 & 98.92/66.67 & 99.14/54.01 & 95.94/32.15 \\
LoRA & 4:1 & 0.0   & 99.17/41.04 & 99.10/42.44 & 97.93/17.81 & 98.69/18.56 & 99.56/84.54 \\
\midrule
LoRA & 4:1 & 0.1  & 99.08/75.44 & 99.13/77.92 & 98.07/8.48 & 98.70/8.33 & 99.69/86.04 \\
LoRA & 4:1 & 0.25 & 99.43/80.33 & 99.18/79.73 & 98.25/42.72 & 98.98/47.17 & 99.65/83.64 \\
LoRA & 4:1 & 0.5  & 99.35/44.50 & 98.98/34.70 & 97.84/14.18 & 97.84/16.30 & 99.68/82.22 \\
\midrule
Frozen & 4:1 & 0.25 & 99.06/64.16 & 98.25/24.70 & 95.11/24.70 & 97.50/23.42 & 99.15/79.90 \\
FT last 2 & 4:1 & 0.25 & 98.86/57.11 & 97.50/22.53 & 94.28/14.31 & 97.08/16.18 & 99.01/79.35 \\
FT last 4 & 4:1 & 0.25 & 99.43/73.64 & 99.05/18.87 & 97.20/28.36 & 98.35/21.27 & 98.89/41.44 \\
\bottomrule
\end{tabular}
}
\end{table}

%% file: tables/exp_eth_recon_full.tex
	\centering
	\captionof{table}{
	\textbf{ETH3D depth consistency benchmark.} Average is the equal-weight mean of the displayed Easy and Hard results. Bold and underline mark the best and second-best results in each column, respectively.
	}
	\label{tab:eth3d-depth-consistency-average-full}
	\footnotesize
	\setlength{\tabcolsep}{2pt}
	\renewcommand{\arraystretch}{0.96}
	\begin{tabular*}{\linewidth}{@{}l@{\extracolsep{\fill}}rrrrrrrrr@{}}
		\toprule
		\multirow{2}{*}{Grouping scorer} & \multicolumn{3}{c}{Easy} & \multicolumn{3}{c}{Hard} & \multicolumn{3}{c}{Average} \\
		\cmidrule(lr){2-4}\cmidrule(lr){5-7}\cmidrule(lr){8-10}
		& MVS$\uparrow$ & AbsRel$\downarrow$ & $\delta_1$(\%)$\uparrow$
		& MVS$\uparrow$ & AbsRel$\downarrow$ & $\delta_1$(\%)$\uparrow$
		& MVS$\uparrow$ & AbsRel$\downarrow$ & $\delta_1$(\%)$\uparrow$ \\
		\midrule
		RobustPi3~\citep{han2025emergent}& 0.230 & 0.287 & 73.6 & 0.133 & 0.394 & 65.9 & 0.1815 & 0.3405 & 69.75 \\
		RobustVGGT & \textbf{0.290} & \textbf{0.239} & \textbf{78.8} & \underline{0.157} & \underline{0.393} & \textbf{71.0} & \textbf{0.2235} & \underline{0.3160} & \textbf{74.90} \\
		GLUEMAP~\citep{pan2026gluemap} & 0.127 & 0.549 & 61.2 & 0.078 & 0.586 & 59.9 & 0.1025 & 0.5675 & 60.55 \\
		\textbf{\algname{} (ours)} & \underline{0.271} & \underline{0.264} & \underline{77.5} & \textbf{0.175} & \textbf{0.321} & \underline{70.9} & \underline{0.2230} & \textbf{0.2925} & \underline{74.20} \\
		\bottomrule
	\end{tabular*}

%% file: tables/exp_eth3d_verification_consistency.tex
\begin{table}[!ht]
  \centering
  \caption{
    \textbf{View graph initialization and multi-view depth consistency on ETH3D.}
    Each setting contains 20 fixed episodes with $K$ scenes and four views per scene.
    Easy (clique) requires pairwise within-scene overlap; Hard (connected) requires a connected overlap graph.
    Only the grouping scorer varies: reconstruction uses the same $\pi^3$ weights,
    shared features, and grouped decoding at $224\!\times\!224$.
  }
  \label{tab:eth3d-verification-consistency}
  \small
  \setlength{\tabcolsep}{4pt}
  \renewcommand{\arraystretch}{1.06}
  \begin{tabular*}{\linewidth}{@{\extracolsep{\fill}}clrrrrrr@{}}
    \toprule
    \multirow{2}{*}{$K$} & \multirow{2}{*}{Grouping scorer}
      & \multicolumn{3}{c}{View graph initialization (\%)}
      & \multicolumn{3}{c}{Depth consistency} \\
    \cmidrule(lr){3-5}\cmidrule(lr){6-8}
      & & AP$\uparrow$ & MR$\uparrow$ & F1$\uparrow$
        & MVS$\uparrow$ & AbsRel$\downarrow$ & $\delta_1$(\%)$\uparrow$ \\
    \midrule
    \multicolumn{8}{@{}l}{\textit{(a) Easy (clique)}} \\
    \addlinespace[2pt]
    \multirow{4}{*}{2} & RobustPi3 & 99.74 & 95.42 & 93.57 & 0.350 & 0.189 & 83.1 \\
     & RobustVGGT & 99.86 & 89.58 & 70.62 & \textbf{0.388} & \textbf{0.181} & \textbf{84.2} \\
     & GLUEMAP & 97.99 & 93.75 & 79.86 & 0.216 & 0.360 & 72.4 \\
     & \textbf{\algname{} (ours)} & \textbf{99.92} & \textbf{99.17} & \textbf{98.09} & 0.339 & 0.191 & 82.8 \\
    \addlinespace[4pt]
    \multirow{4}{*}{3} & RobustPi3 & 97.03 & 63.06 & 89.56 & 0.214 & 0.272 & 75.4 \\
     & RobustVGGT & 92.22 & 33.33 & 60.48 & \textbf{0.293} & \textbf{0.247} & \textbf{80.4} \\
     & GLUEMAP & 96.80 & 93.33 & 66.04 & 0.109 & 0.570 & 61.2 \\
     & \textbf{\algname{} (ours)} & \textbf{99.22} & \textbf{94.17} & \textbf{94.21} & 0.241 & 0.330 & 75.7 \\
    \addlinespace[4pt]
    \multirow{4}{*}{4} & RobustPi3 & 96.31 & 49.17 & 82.02 & 0.127 & 0.401 & 62.3 \\
     & RobustVGGT & 76.03 & 25.00 & 47.98 & 0.189 & 0.290 & 71.7 \\
     & GLUEMAP & 96.06 & 92.92 & 57.78 & 0.057 & 0.718 & 50.0 \\
     & \textbf{\algname{} (ours)} & \textbf{99.54} & \textbf{94.58} & \textbf{98.34} & \textbf{0.234} & \textbf{0.270} & \textbf{74.0} \\
    \midrule
    \multicolumn{8}{@{}l}{\textit{(b) Hard (connected)}} \\
    \addlinespace[2pt]
    \multirow{4}{*}{2} & RobustPi3 & \textbf{99.95} & \textbf{97.39} & 96.27 & \textbf{0.222} & \textbf{0.277} & \textbf{77.7} \\
     & RobustVGGT & 99.47 & 94.77 & 68.10 & 0.194 & 0.282 & 76.3 \\
     & GLUEMAP & 97.54 & 92.16 & 75.25 & 0.132 & 0.350 & 73.3 \\
     & \textbf{\algname{} (ours)} & 99.92 & 94.77 & \textbf{99.03} & 0.213 & 0.281 & 77.1 \\
    \addlinespace[4pt]
    \multirow{4}{*}{3} & RobustPi3 & 97.79 & 52.16 & 85.33 & 0.115 & 0.411 & 64.5 \\
     & RobustVGGT & 79.00 & 31.47 & 55.08 & 0.148 & \textbf{0.326} & \textbf{71.1} \\
     & GLUEMAP & 96.19 & \textbf{92.67} & 57.47 & 0.064 & 0.593 & 59.1 \\
     & \textbf{\algname{} (ours)} & \textbf{99.34} & 87.50 & \textbf{96.51} & \textbf{0.165} & 0.330 & 70.8 \\
    \addlinespace[4pt]
    \multirow{4}{*}{4} & RobustPi3 & 95.75 & 35.06 & 74.57 & 0.061 & 0.493 & 55.6 \\
     & RobustVGGT & 72.91 & 25.65 & 50.42 & 0.128 & 0.571 & \textbf{65.7} \\
     & GLUEMAP & 96.57 & \textbf{94.16} & 49.55 & 0.037 & 0.816 & 47.2 \\
     & \textbf{\algname{} (ours)} & \textbf{98.34} & 78.25 & \textbf{91.92} & \textbf{0.145} & \textbf{0.352} & 64.6 \\
    \bottomrule
  \end{tabular*}
\end{table}

%% file: tables/appendix_pose.tex
\begin{table}[!ht]
  \centering
  \caption{\textbf{Full pose estimation accuracy.}
  Supplement to Table~\ref{tab:pose_estimation}, using the same strict score
  $>0.5$ verification rule. AP-RPE and SA-RPE are defined in
  Appendix~\ref{app:pose-estimation}. Bold and underline mark the best and
  second-best results per dataset and metric.}
  \label{tab:pose-system-aligned-full}
  \small
  \setlength{\tabcolsep}{5pt}
  \resizebox{\linewidth}{!}{%
  \begin{tabular}{@{}lrrrrrr@{}}
    \toprule
    \multirow{2}{*}{Method}
    & \multicolumn{2}{c}{AP-RPE median $\downarrow$}
    & \multicolumn{2}{c}{SA-RPE mean $\downarrow$}
    & \multicolumn{2}{c}{System accuracy (\%) $\uparrow$} \\
    \cmidrule(lr){2-3}\cmidrule(lr){4-5}\cmidrule(l){6-7}
    & Rot. ($^\circ$) & Dir. ($^\circ$)
    & Rot. ($^\circ$) & Dir. ($^\circ$)
    & $10^\circ$/1\,m & $20^\circ$/5\,m \\
    \midrule
    \multicolumn{7}{@{}l}{\textit{ETH3D}} \\
    Pi3~\citep{wang2025pi3} & 4.67 & 16.04 & 99.60 & 107.58 & 27.55 & 41.04 \\
    VGGT~\citep{wang2025vggt} & 14.76 & 43.25 & 117.39 & 126.79 & 8.43 & 22.99 \\
    RobustPi3~\citep{han2025emergent} & \textbf{4.46} & \underline{15.52} & \underline{20.70} & \underline{28.53} & \underline{73.08} & \underline{85.71} \\
    RobustVGGT~\citep{han2025emergent} & 8.57 & 25.90 & 61.44 & 65.45 & 54.90 & 62.82 \\
    GLUEMAP~\citep{pan2026gluemap} & 4.52 & \textbf{15.49} & 45.05 & 52.33 & 60.00 & 72.04 \\
    \algname (Ours) & \underline{4.48} & 15.56 & \textbf{15.39} & \textbf{23.02} & \textbf{76.19} & \textbf{88.09} \\
    \midrule
    \multicolumn{7}{@{}l}{\textit{OpenNavMap}} \\
    Pi3~\citep{wang2025pi3} & 3.01 & 5.84 & 107.10 & 111.51 & 19.87 & 31.17 \\
    VGGT~\citep{wang2025vggt} & 5.83 & 15.10 & 109.70 & 117.43 & 8.89 & 28.60 \\
    RobustPi3~\citep{han2025emergent} & \textbf{2.65} & \underline{5.41} & 54.27 & 57.74 & 60.92 & 68.36 \\
    RobustVGGT~\citep{han2025emergent} & 5.50 & 14.66 & 45.07 & 51.98 & 48.13 & 70.63 \\
    GLUEMAP~\citep{pan2026gluemap} & \underline{2.74} & \textbf{5.21} & \textbf{32.18} & \textbf{36.71} & \textbf{67.83} & \textbf{79.72} \\
    \algname (Ours) & 2.81 & 6.46 & \underline{41.18} & \underline{45.22} & \underline{67.65} & \underline{75.86} \\
    \midrule
    \multicolumn{7}{@{}l}{\textit{360Loc atrium}} \\
    Pi3~\citep{wang2025pi3} & 12.18 & 21.11 & 125.04 & 126.58 & 1.86 & 16.86 \\
    VGGT~\citep{wang2025vggt} & 15.26 & 31.13 & 127.35 & 134.89 & 0.25 & 12.06 \\
    RobustPi3~\citep{han2025emergent} & \textbf{5.52} & \textbf{10.93} & 60.97 & 62.61 & \textbf{54.07} & 64.71 \\
    RobustVGGT~\citep{han2025emergent} & 12.22 & 27.45 & 57.62 & 61.12 & 49.75 & 62.75 \\
    GLUEMAP~\citep{pan2026gluemap} & \underline{7.95} & \underline{13.75} & \underline{54.09} & \underline{55.47} & 52.79 & \underline{66.37} \\
    \algname (Ours) & 10.48 & 18.76 & \textbf{38.46} & \textbf{39.56} & \underline{53.82} & \textbf{72.70} \\
    \bottomrule
  \end{tabular}%
  }
\end{table}

%% file: tables/appendix_opennavmap.tex
\begin{table}[!ht]
  \centering
  \small
  \caption{Cross-session localization on OpenNavMap. CosPlace retrieves the top eight candidates per query, which each method reranks using its saved verification scores. Recall@K is the fraction of answerable queries with at least one correct result among the top K; correctness uses ground-truth camera-center distance $\leq 25$\,m and relative orientation $\leq 40^\circ$. All entries are percentages. The CosPlace R@8 columns show the retrieval ceiling shared by every reranker. OpenNavMap is the equal-weight mean of Region A and Region B.}
  \label{tab:opennavmap-reranking-pose-full}
  \renewcommand{\arraystretch}{1.2}
  \resizebox{\linewidth}{!}{%
  \begin{tabular}{@{}l*{12}{c}@{}}
    \toprule
    & \multicolumn{4}{c}{Region A}
    & \multicolumn{4}{c}{Region B}
    & \multicolumn{4}{c}{OpenNavMap (macro)} \\
    \cmidrule(lr){2-5}\cmidrule(lr){6-9}\cmidrule(l){10-13}
    Ranking & R@1 & R@3 & R@5 & R@8
            & R@1 & R@3 & R@5 & R@8
            & R@1 & R@3 & R@5 & R@8 \\
    \midrule
    CosPlace
      & 63.93 & 69.48 & 73.01 & 76.04
      & 66.05 & 71.60 & 73.95 & 76.49
      & 64.99 & 70.54 & 73.48 & 76.26 \\
    RobustPi3
      & 65.83 & 73.77 & 75.28 & --
      & 70.35 & 74.30 & 75.56 & --
      & 68.09 & 74.04 & 75.42 & -- \\
    RobustVGGT
      & 71.25 & 74.65 & 75.28 & --
      & 71.35 & 74.21 & 75.26 & --
      & 71.30 & 74.43 & 75.27 & -- \\
    GLUEMAP$^\dagger$
      & 72.26 & 75.41 & 75.91 & --
      & 71.33 & 73.74 & 75.19 & --
      & 71.79 & 74.58 & 75.55 & -- \\
    \algname (Ours)
      & \textbf{74.15} & \textbf{75.54} & \textbf{76.04} & --
      & \textbf{72.88} & \textbf{75.00} & \textbf{75.79} & --
      & \textbf{73.52} & \textbf{75.27} & \textbf{75.92} & -- \\
    \bottomrule
  \end{tabular}%
  }
\end{table}

%% file: tables/appendix_runtime.tex
\begin{table}[H]
  \centering
  \caption{
    \textbf{OpenNavMap runtime and peak GPU memory: all predicted groups.}
    Each batch contains nine images and 36 unordered pairs.
    Each method uses its own verification scores with the strict
    score $>0.5$ connected-component rule.
    Equivalent pairs/s is 36 times batch/s.
    Average and maximum peak memory subtract each run's measured idle baseline
    and use decimal GB. Best and second-best results are bold and underlined,
    respectively.
  }
  \label{tab:opennavmap-runtime-all-full}
  \resizebox{\linewidth}{!}{%
  \begin{tabular}{@{}lrrrrr@{}}
    \toprule
    Method & s/batch $\downarrow$ & batch/s $\uparrow$
      & Equivalent pairs/s $\uparrow$ & Avg peak GB $\downarrow$ & Max peak GB $\downarrow$ \\
    \midrule
    RobustPi3 & \underline{0.218} & \underline{4.588} & \underline{165.180} & \underline{4.727} & \underline{4.727} \\
    RobustVGGT & 0.693 & 1.442 & 51.912 & 10.994 & 11.065 \\
    GLUEMAP & 4.663 & 0.214 & 7.721 & \textbf{4.633} & \textbf{4.635} \\
    \algname (Ours) & \textbf{0.159} & \textbf{6.305} & \textbf{226.995} & 8.087 & 8.087 \\
    \bottomrule
  \end{tabular}%
  }
\end{table}

\begin{table}[H]
  \centering
  \caption{
    \textbf{OpenNavMap runtime and peak GPU memory: query group only.}
    The output scope is restricted to the connected component containing query
    image 0, using the same nine-image, 36-pair batches and strict score $>0.5$
    rule as Table~\ref{tab:opennavmap-runtime-all-full}.
    Equivalent pairs/s is 36 times batch/s.
    Average and maximum peak memory subtract each run's measured idle baseline
    and use decimal GB. Best and second-best results are bold and underlined,
    respectively.
  }
  \label{tab:opennavmap-runtime-query-full}
  \resizebox{\linewidth}{!}{%
  \begin{tabular}{@{}lrrrrr@{}}
    \toprule
    Method & s/batch $\downarrow$ & batch/s $\uparrow$
      & Equivalent pairs/s $\uparrow$ & Avg peak GB $\downarrow$ & Max peak GB $\downarrow$ \\
    \midrule
    RobustPi3 & \underline{0.158} & \underline{6.327} & \underline{227.773} & \underline{4.727} & \underline{4.727} \\
    RobustVGGT & 0.421 & 2.376 & 85.550 & 10.995 & 11.065 \\
    GLUEMAP & 4.564 & 0.219 & 7.887 & \textbf{4.634} & \textbf{4.637} \\
    \algname (Ours) & \textbf{0.127} & \textbf{7.851} & \textbf{282.621} & 8.087 & 8.087 \\
    \bottomrule
  \end{tabular}%
  }
\end{table}

%% file: tables/appendix_pairwise.tex
\begin{table}[H]
	\centering
	\caption{
	\textbf{Pairwise verification.}
	We evaluate on VisymScenes~\citep{xiangli2025doppelgangers} and GV-Bench~\citep{yu2024gv} with average precision (AP) and maximum recall at 100\% precision (MR), both expressed as percentages. Avg.\ reports the mean of each metric across the two benchmarks.
	The first three methods are not trained on VisymScenes; the remaining methods are. Bold marks the best result per column.
	}
	\label{tab:pairwise-verify}
	\resizebox{\linewidth}{!}{
		\begin{tabular}{lcccccc}
			\toprule
			\multirow{2}{*}{Method} & \multicolumn{2}{c}{VisymScenes} & \multicolumn{2}{c}{GV-Bench} & \multicolumn{2}{c}{Avg.}  \\
			& AP $\uparrow$ & MR $\uparrow$ & AP $\uparrow$ & MR $\uparrow$ & AP $\uparrow$ & MR $\uparrow$ \\
			\midrule
			MegaLoc~\citep{Berton_2025_MegaLoc} & 91.17 & 26.42 & \tb{91.74} & \tb{32.38} & 91.46 & 29.40 \\
			RoMAv2~\citep{edstedt2025romav2} & 92.75 & 2.45 & 84.63 & 19.49 & 88.69 & 10.97 \\
			RobustVGGT~\citep{han2025emergent} & 93.57 & 18.81 & 52.13 & 2.14 & 72.85 & 10.48 \\
			\midrule
			Doppelgangers~\citep{cai2023doppelgangers}  & 97.89 & 38.21 & 73.95 & 17.28 & 85.92 & 27.75 \\
			Doppelgangers++~\citep{xiangli2025doppelgangers}  & 98.42 & 41.76 & 89.70 & 25.75 & \tb{94.06} & 33.76 \\
			\algname (FT last 2) & \textbf{99.05} & \textbf{74.34} & 81.45 & 11.18 & 90.25 & \tb{42.76} \\
			\algname (LoRA) & 98.72 & 60.92 & 85.17 & 15.79 & 91.95 & 38.36 \\
            \bottomrule
		\end{tabular}
    }
\end{table}

\begin{table}[H]
	\centering
	\caption{
	\textbf{Pairwise verification on GV-Bench (per condition).} We report AP and MR (\%) across the six GV-Bench conditions; Avg.\ is the mean over conditions. Best result per column in \textbf{bold}.
	}
	\label{tab:pairwise-gvbench}
	\setlength{\tabcolsep}{4pt}
	\resizebox{\linewidth}{!}{
		\begin{tabular}{lcccccccccccccc}
			\toprule
			\multirow{2}{*}{Method} & \multicolumn{2}{c}{Day} & \multicolumn{2}{c}{Night} & \multicolumn{2}{c}{Weather} & \multicolumn{2}{c}{Season} & \multicolumn{2}{c}{Nordland} & \multicolumn{2}{c}{Uacampus} & \multicolumn{2}{c}{Avg.} \\
			& AP $\uparrow$ & MR $\uparrow$ & AP $\uparrow$ & MR $\uparrow$ & AP $\uparrow$ & MR $\uparrow$ & AP $\uparrow$ & MR $\uparrow$ & AP $\uparrow$ & MR $\uparrow$ & AP $\uparrow$ & MR $\uparrow$ & AP $\uparrow$ & MR $\uparrow$ \\
			\midrule
			MegaLoc & 99.64 & \tb{57.84} & 97.81 & \tb{28.58} & 99.77 & \tb{49.26} & \tb{99.81} & \tb{48.72} & \tb{77.32} & \tb{3.15} & 76.09 & 6.72 & \tb{91.74} & \tb{32.38} \\
			RoMAv2 & \tb{99.78} & 50.31 & \tb{97.89} & 13.47 & 99.82 & 14.89 & 99.75 & 7.33 & 29.30 & 0.11 & \tb{81.22} & \tb{30.85} & 84.63 & 19.49 \\
			\algname (FT last 2) & 98.53 & 32.39 & 80.00 & 1.63 & 99.57 & 13.68 & 99.34 & 16.20 & 46.21 & 0.17 & 65.06 & 2.99 & 81.45 & 11.18 \\
			\algname (LoRA) & 99.20 & 40.63 & 87.15 & 3.79 & \tb{99.83} & 18.90 & 99.46 & 27.90 & 52.19 & 1.30 & 73.17 & 2.24 & 85.17 & 15.79 \\
			RobustVGGT & 78.74 & 7.66 & 36.19 & 0.19 & 94.93 & 1.08 & 88.76 & 0.06 & 6.42 & 0.11 & 7.76 & 3.73 & 52.13 & 2.14 \\
			\bottomrule
		\end{tabular}
	}
\end{table}